\pdfoutput=1

\documentclass[11pt]{article}

\usepackage{acl}

\usepackage{times}
\usepackage{latexsym}

\usepackage[T1]{fontenc}

\usepackage[utf8]{inputenc}

\usepackage{microtype}

\usepackage{graphicx}
\usepackage{algorithm}
\usepackage{algorithmic}
\usepackage{amssymb}
\usepackage{xcolor}         
\usepackage{subfigure}
\usepackage{amsmath}
\usepackage{amsthm}
\usepackage{multirow}
\usepackage{diagbox}
\usepackage{enumitem}
\usepackage{bm}
\usepackage{booktabs}

\newcommand{\bx}{\mathbf{x}}

\newcommand{\bt}{\mathbf{t}}
\newcommand{\by}{\mathbf{y}}
\newcommand{\bz}{\mathbf{z}}
\newcommand{\btau}{\boldsymbol{\tau}}

\newcommand{\bh}{\mathbf{h}}

\newcommand{\mL}{\mathcal{L}}

\title{Towards Unifying the Label Space for Aspect- and Sentence-based Sentiment Analysis}

\author{Yiming Zhang \\ Zhejiang University \\ \texttt{yimingz@zju.edu.cn} \\
        {\bf Sai Wu} \\ Zhejiang University \\ \texttt{wusai@zju.edu.cn} \\
        \And  Min Zhang \\ Zhejiang University \\ \texttt{min\_zhang@zju.edu.cn} \\
        {\bf Junbo Zhao (Jake)} \\ Zhejiang University \\ \texttt{j.zhao@zju.edu.cn}}

\begin{document}
\maketitle
\begin{abstract}
The aspect-based sentiment analysis (ABSA) is a fine-grained task that aims to determine the sentiment polarity towards targeted aspect terms occurring in the sentence.
The development of the ABSA task is very much hindered by the lack of annotated data.
To tackle this, the prior works have studied the possibility of utilizing the sentiment analysis (SA) datasets to assist in training the ABSA model, primarily via pretraining or multi-task learning.
In this article, we follow this line, and for the first time, we manage to apply the Pseudo-Label (PL) method to merge the two homogeneous tasks. 
While it seems straightforward to use generated pseudo labels to handle this case of label granularity unification for two highly related tasks, we identify its major challenge in this paper and propose a novel framework, dubbed as Dual-granularity Pseudo Labeling (DPL).
Further, similar to PL, we regard the DPL as a general framework capable of combining other prior methods in the literature ~\cite{rietzler2019adapt,bai2020investigating}.
Through extensive experiments, DPL has achieved state-of-the-art performance on standard benchmarks surpassing the prior work significantly ~\cite{liu2021pre}.

\end{abstract}

\section{Introduction}



\subsection{Aspect-based Sentiment Analysis}
\begin{figure}
  \centering
  \includegraphics[width=0.8\linewidth]{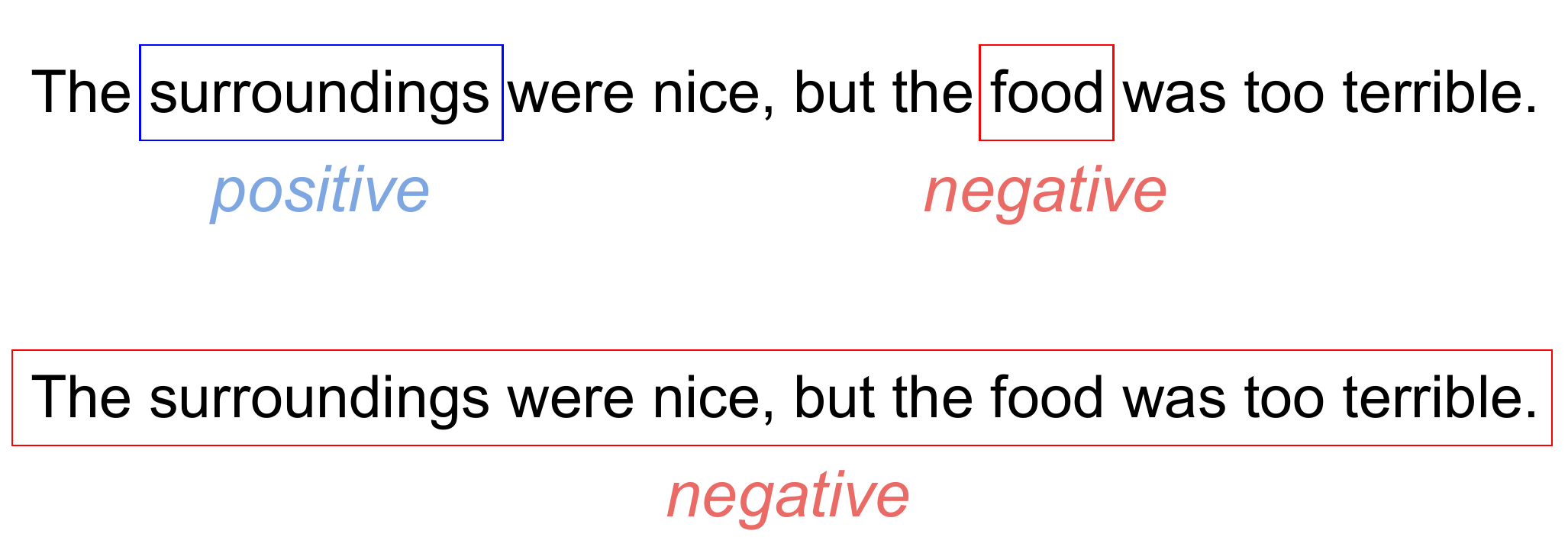}
  \caption{Sentiment Analysis (SA) and Aspect-based Sentiment Analysis (ABSA). The sample on the above is the ABSA task, while the sample on the bottom is the SA task. Both tasks aim at analyzing the sentiments carried by the objects in the box.}
  \label{fig:sa-absa}
\end{figure}

The aspect-based sentiment analysis (ABSA) task aims to recognize the sentiment polarities centered on the considered aspect terms occurring in the sentence.
The establishment of the ABSA task echoes the long-standing literature of conventional sentence-level sentiment analysis (SA).
For instance, as shown in Figure ~\ref{fig:sa-absa}, a normal ABSA data annotation tags sentiment score on specific aspect terms in the sentence, like ``surroundings'' as positive and ``food'' as negative. Meanwhile, in the conventional sentence-based sentiment analysis, the whole sentence is labeled as negative at a coarser granularity.



Due to its much finer granularity, the annotation cost is significantly higher than its conventional counterpart. Essentially, many of the existing SA datasets ~\cite{he2018exploiting} can be crawled and curated straightforwardly from the review websites such as Amazon\footnote{\url{https://www.amazon.com/}} or Yelp\footnote{\url{https://www.yelp.com/}}. The five-star rating system comes in handy to accomplish the annotation. Thus, the SA datasets are often presented at a large scale.
By contrast, the ABSA annotation has no such ``free lunch''. It has to require human annotators to participate. Coupling with its higher complexity on labeling, the ABSA datasets are ubiquitously at considerably smaller scales ~\cite{pontikisemeval,he2018exploiting,yu2021cross}.
To this date, the available datasets for conventional sentiment analysis are generally larger to several orders of magnitude than the ABSA. 
\begin{figure*}
  \centering
  \includegraphics[width=12cm]{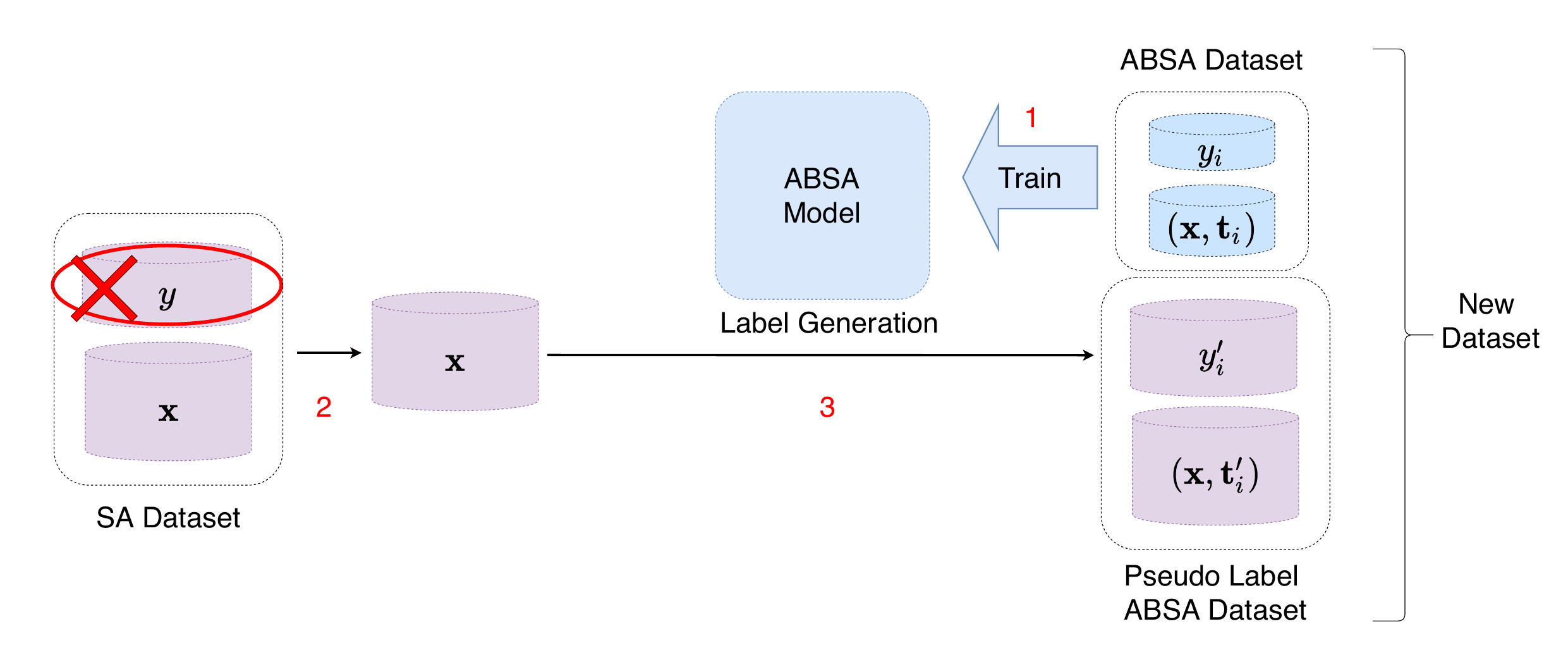} 
  \caption{Dataset Generation in the Pseudo-Label (PL) Method.
  This figure shows a pipeline of the traditional Pseudo-Label method.
  $\bx$ is the input data, a sentence in the SA dataset, while $y$ is the sentiment carried by a sentence. $\bt_i$ indicates the position of an aspect term in a sentence, and $y_i$ is the label for that aspect term. $\bt'_i$ and $y'_i$ are pseudo labels generated by the ABSA model.
  As we can see, in the PL method, the sentence sentiment labels are dropped, and the SA dataset is regarded as an unlabeled dataset.}
  \label{fig:pl}
\end{figure*}

For instance, the commonly used ABSA benchmark SemEval 2014 task 4 has less than 5000 samples ~\cite{pontikisemeval}, while there are 4,000,000 sentences in the Amazon Review dataset\footnote{\url{https://www.kaggle.com/bittlingmayer/amazonreviews}} for SA.
Due to the similarity between the SA task and the ABSA task, it is natural to use SA datasets as auxiliary datasets for the ABSA task ~\cite{he2018exploiting}. 
Most, if not all,  previous work has focused on pretraining and multi-task learning methods ~\cite{he2018exploiting,he2019interactive}.
In this paper, we first take the Pseudo-Label method to utilize the SA datasets to solve the challenge faced by the ABSA task.

\subsection{Pseudo-Label}
The family of Pseudo-Label methods has had wide success in multiple fields ~\cite{pham2020meta,ge2020mutual,mallis2020unsupervised,zoph2020rethinking,he2019revisiting}.
The core of this family is to ``trust'' the generated fake labels by running the unlabeled samples through a teacher network that is trained by using the limited number of labeled samples.
The generated labeled samples are then combined with the original set of supervised datasets and fed to the final model training.

In this article, our core mission is to incorporate the large-scale datasets into the sentiment analysis with the targeted ABSA task.
While there have been works on this line, such as ~\citet{he2018exploiting} and ~\citet{he2019interactive}, exploring the Pseudo-Label methods has been very much untapped. Indeed, a very straightforward technological solution is depicted in Figure ~\ref{fig:pl}. 
One can apply the traditional Pseudo-Label method to generate a bunch of pseudo-aspect-based sentiment labels from the SA or even the unlabeled datasets. 
However, a consequence of this is the total abandonment and waste of the provided coarse-grained labels.
While seemingly acceptable, we argue that due to the homogeneous root for the ABSA and SA tasks, the under-exploiting of the sentence-level coarse-grained sentiment labels is sub-optimal.
It will be unnecessary if the traditional framework throws away the coarser-grained labels containing finer-grained task-relevant information.
We argue that the Pseudo-Label family of approaches is limited to fit a \emph{uniform granularity} situation. They ought to evolve and further adapt to the discrepancy of granularity in the label space.

\subsection{Dual-granularity Pseudo Labels}
To solve the aforementioned problem, we propose the Dual-granularity Pseudo Labeling framework (DPL). 
In essence, the DPL augments the original PL framework and is capable of leveraging the labels drawn from both granularities.
Briefly, the DPL relies on two teacher models obtained from datasets from both granularities, respectively, then generates pseudo labels for both sides. 
As a result, datasets from both granularity levels can be merged into a whole, with every sentence sample being tagged by both finer- and coarser-set of labels. 
To facilitate the employment of both sets of labels, we set a few standard conditions as the design principle of DPL. Slightly more concretely, DPL establishes two separate pathways leading to prediction for both granularities. Together, the two pathways interact in the representation space and ideally may possess controlled information flow that respectively corresponds and only correspond to the considered granularity.
We incorporate an adversarial module to accomplish this functionality.

On the widely used benchmarks, SemEval 2014 task 4 subtask 2 ~\cite{pontikisemeval}, the DPL method significantly surpasses the current state-of-the-art method. We deem our simple but effective framework DPL pioneering a bi-granularity level dataset merging. 
In what follows, we empirically validate that DPL is a framework that can be seamlessly combined with the previous pre-training or multi-task learning methods in terms of ABSA and SA dataset merging.

To sum up, we make the following contributions in this paper:
\begin{enumerate}
    \item Among those works to solve the lack of labeled data in the ABSA task, we pioneer to adopt and enhance a pseudo-label framework to leverage the coarser-grained SA labels.
    \item We propose a novel general framework called Dual-granularity Pseudo Labels (DPL). Just like the vanilla PL method, the DPL is established as a general framework. We validate that DPL is also compatible with previous work on this line, such as pre-training or multi-task learning (MTL). DPL has achieved excellent performances on the standardized ABSA benchmarks such as SemEval 2014, which significantly outperforms the prior works.
\end{enumerate}

\section{Related Works}
\subsection{Aspect-based Sentiment Analysis (ABSA)}
ABSA is a finer-grained task of Sentiment Analysis (SA). 
It is a pipeline task, including aspect term extraction and aspect term sentiment classification. 
Aspect term sentiment classification is the true target task in this paper.
For convenience, we use ABSA to refer to this task in the remaining parts.

Like other application tracks in NLP, the family of neural network models has wide successes in this task ~\cite{jiang2011target,vo2015target,zhang2016gated,ma2017interactive,li2018hierarchical,wang2018target,huang2018aspect,song2019attentional}.
~\citet{wang2016attention} introduce attention mechanism into an LSTM to model the inter-dependence between sentence and aspect term. 
~\citet{tang2016aspect} apply Memory Networks in this task.

Syntax-based models have also been explored widely in this domain ~\cite{dong2014adaptive,tai2015improved,nguyen2015phrasernn,liu2020jointly,li2021dual,pang2021dynamic}.
\citet{sun2019aspect} and \citet{zhang2019aspect} introduced graph convolution networks (GCN) to leverage the structured information from the dependency tree.
\citet{huang2019syntax} used graph attention networks (GAT) to improve the performance.
\citet{bai2020investigating} and \citet{wang2020relational} took the syntax relations as edge features and introduced them into the Relational Graph Attention Network (RGAT). 

In addition, pretrained language models like BERT ~\cite{devlin2018bert} have greatly promoted the development of ABSA ~\cite{li2018hierarchical,gao2019target,song2019attentional,rietzler2019adapt,yang2019multi}.

\subsection{Using Extra Dataset for ABSA}
Due to the dataset scale challenge of the ABSA task, there have been some methods exploring how to utilize the auxiliary dataset.

Some of them ~\cite{xu2019bert,rietzler2019adapt,yu2021cross} achieve decent ABSA performance by post-processing or fine-tuning BERT ~\cite{devlin2018bert} with an additional unlabeled dataset.
For these methods, we argue that the cost of computation is too high. Moreover, DPL does not conflict with it and can accommodate the results of these works. We take \citet{rietzler2019adapt}'s work as an example for comparison in experiments.

The others ~\cite{he2018exploiting,he2019interactive,chen2019transfer,liang2020iterative,yang2019multi,oh2021deep,yu2021making,yan2021unified} utilize some labeled datasets and propose (later extend) a framework applying multitask learning methods. These auxiliary labeled datasets mainly include the sentiment analysis (SA) task and other subtasks of ABSA, such as Aspect Term Extraction, Opinion Term Extraction, and so on ~\cite{yan2021unified}. 
DPL is more similar to these methods, using labeled datasets. 
However, we argue that the datasets of other subtasks can't solve the problem of the high annotation cost.
Thus, DPL utilizes the SA task as auxiliary datasets and is the first to apply the PL method to this problem.

\subsection{Pseudo-Label}
Pseudo-label (PL), often associated with self-training, is a semi-supervised learning method. PL has been utilized and further developed by many studies ~\cite{ge2020mutual,mallis2020unsupervised,zoph2020rethinking,he2019revisiting}. It has been successfully applied in many tasks, such as image classification ~\cite{pham2020meta,xie2020self}, object detection ~\cite{ge2020mutual}, text classification ~\cite{mukherjee2020uncertainty}, Etc.

However, these PL methods are inapplicable under a non-uniform granularity situation; that is, there are massive available coarse-grained datasets for fine-grained tasks. These existing methods can only discard the coarse-grained labels and treat them as unlabeled datasets. Thus, we argue that these PL methods cause loss of information and are definitely unreasonable.

\section{Preliminary}
\subsection{Pseudo-Labels}
The traditional PL method generally involves a labeled set denoted by $D$ and an unlabeled set $D_u$.
A teacher model is trained on $D$ by cross-entropy loss:
\begin{equation}
    \mL(\Theta_{T}) = \sum_{(x,y) \in D} [-\log P_{\Theta_{T}}(y|x)]
    \label{Teacher-Training}
\end{equation}
where $\Theta_{T}$ denotes the parameters of the teacher model.
The cross-entropy loss function is adopted for general classification problems, including image classification, detection, and semantic segmentation~\cite{ge2020mutual,pham2020meta,xie2020self,zoph2020rethinking}. 

In what follows, on the unlabeled dataset $D_u$, one can obtain the corresponding labels via running the unlabeled input through an inference procedure of the teacher model. 
The yielded label set for $D_u$ forms a pseudo-labeled dataset that can later be combined with the original dataset with gold annotations. 
A student model $M_S$ is trained by the newly merged dataset:
\begin{equation}
\begin{aligned}
    \mL(\Theta_{S}) = 
    & \sum_{(x,y) \in D} [-\log P_{\Theta_{S}}(y|x)] + \\
    & \lambda \sum_{(x_u, y') \in D'_u} [-\log P_{\Theta_{S}}(y'|x_u)]
\end{aligned}
\end{equation}
where $y'$ indicates the pseudo label corresponding to the sample $x_u$ generated by the teacher model. $D'_u$ are the pseudo-label augmented version of $D_u$. $\lambda$ is a weighing term.




\section{Dual-granularity Pseudo Labeling}

In short, our work focuses on expanding the traditional PL method to utilize coarse-grained datasets. To achieve this goal, we draw inspiration from the multi-task learning community and augment the PL method with a different modeling pathway. Consequently, we obtain a framework where two separate pathways are trained synergistically targeted at labels of both granularities.

\subsection{Setup}
Our work is based on two datasets, the fine-grained and the coarse-grained datasets in the same domain.
Let us use $D_{\text{fine}}$ and $D_{\text{coarse}}$ to denote two datasets respectively.
For the coarse-grained dataset $D_{\text{coarse}}$, the task is to learn a mapping:
\begin{equation}
    f_{\text{coarse}}(\bx) \to y,
\end{equation}
For the fine-grained dataset $D_{\text{finer}}$, the target mapping is:
\begin{equation}
    f_{\text{fine}}(\bx, \bt_i) \to y_i, i \in \{1,...,m\}
\end{equation}
where $(\bx, y) \in D_{\text{coarse}}$ and $(\bx, \bt_i, y_i) \in D_{\text{fine}}$. 
$\bx$ is the input data, and $y$ is the corresponding label for $\bx$.
$\bt_i \subseteq \bx$. $m$ means that $\bx$ has m sub-parts totally, and $y_i$ is the corresponding label for $\bt_i$.

The traditional PL method is limited to fit a uniform granularity situation.
The first step to resolve this limitation is to merge the coarse-grained dataset with the fine-grained dataset.
Like the traditional PL method, we train a teacher model on one dataset and generate pseudo labels for the other dataset. We repeat this process at two granularities.
Here, we suppose that $\bx_i$ for each $\bx$ in the $D_{\text{coarse}}$ have been extracted.
After pseudo labels generation, two new datasets are generated, donates as $D'_{\text{fine}}$ and $D'_{\text{coarse}}$, and a new dataset $D'$ are merged by these two datasets.
Specifically, 

\begin{equation}
    D' = D'_{\text{fine}} \cup D'_{\text{coarse}},
\end{equation}
where $(\bx, \bt_i, y, y'_i) \in D'_{\text{coarse}}$ and $(\bx, \bt_i, y', y_i) \in D'_{\text{fine}}$.
$y'$ and $y'_i$ are the generated pseudo labels.

Up to now, we get a new dataset with a much larger scale. Our goal translates into obtaining a model trained by the new dataset $D'$ with high performance on the fine-grained task.
In other words, compared with the traditional PL method, 
the key problem is: how to utilize the coarse-grained labels to improve the model's performance on the fine-grained task.

\begin{figure*}
  \centering
  \includegraphics[width=13cm]{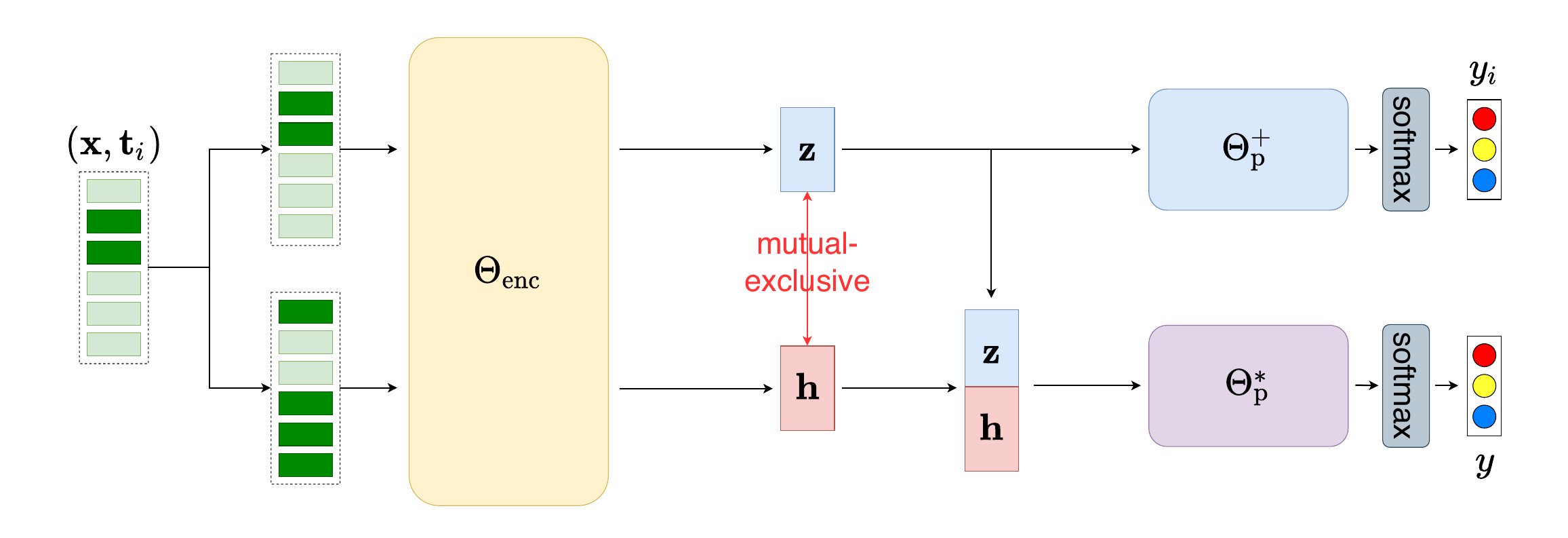} 
  \caption{The Model for the DPL Framework. $(\bx, \bt_i)$ is the input data. $\bt_i$ indicates the aspect terms, which are painted by the dark green. We first generate $(\bx, \bt_i)$ and $(\bx, \mathbf{1}-\bt_i)$ as the input of the upper and lower pathways, respectively. In this case, $\bt_i = (0,1,1,0,0,0)$ and $\mathbf{1}-\bt_i = (1,0,0,1,1,1)$. ``$\Theta_{\text{enc}}$'' is an encoder that outputs $\bz$ and $\bh$. ``$\Theta^+_{\text{p}}$'' is a predictor for the fine-grained task, and ``$\Theta^*_{\text{p}}$'' is a predictor for the coarse-grained task. Correspondingly, $y_i$ is the prediction for the fine-grained task, and $y$ is the prediction for the coarse-grained task. ``mutual-exclusive'' means the information carried by $\bz$ and $\bh$ has little overlap.}
  \label{fig:model}
\end{figure*}

\subsection{DPL Skeleton}

As we mentioned, the core challenge for adapting the vanilla PL method is to utilize coarse-grained labels.
As displayed in Figure ~\ref{fig:model}, we set dual pathways corresponding to each granularity. Both pathways are finished by setting a proper softmax-based classifier.
Using $\bz$ and $\bh$ to denote the internal representation vectors for both pathways, we decompose the design philosophy of DPL by the following three conditions:
\begin{itemize}
    \item
    $\bz$ carries adequate information to determine the label at the fine-grained level. More formally, there exists a function $f_{\Theta^+_\text{p}}$ in the overall functional space that is able to map the $\bz$ to $y_i$.
    \item The union set of $\bh$ and $\bz$ is capable of determining the label at the coarse level.
    There exists another function $f_{\Theta^*_\text{p}}$ in the overall functional space that is sufficient to map the $[\bz \circ \bh]$ to $y$.
    \item $\bh$ and $\bz$ are mutually exclusive in terms of the carried information. That means we cannot train a function $f_{\Theta^*_\text{p}}$ to map $\bh$ to $y_i$, due to the lack of information contributed from $\bz$.
\end{itemize}


The main rationale behind these three conditions may include but is not limited to: (i)-the information passing through the pathway with $\bz$ is only required in the fine-grained task; (ii)-the other information needed by the coarse-grained task passes through the pathway with $\bh$; (iii)-the prediction at coarse-grained level is based on the concatenation of $\bh$ and $\bz$, while either of them is insufficient to accomplish the prediction of coarse-grained labels.


In order to satisfy the model to these three conditions, our loss function consists of three terms.
Among them, the two terms are the classification loss terms for the fine- and coarse-grained tasks, respectively, fulfilling conditions 1\&2.
For condition 3,  we draw inspiration from adversarial training ~\cite{lample2017fader} to reduce the fine-grained task-relevant information carried by $\bh$. 


\subsubsection{Fine- and Coarse-grained Tasks}
As shown in Figure~\ref{fig:model}, the model consists of an encoder, $\Theta_{\text{enc}}$, together with two predictors, $\Theta^*_{\text{p}}$  and $\Theta^+_{\text{p}}$. 
In particular, $\Theta_{\text{enc}}$ encodes each input data $(\bx, \bt_i)$ into two intermediate results, $\bz$ and $\bh$. 
In the figure, the top line with $\bz$ is the pathway for the fine-grained task-relevant information flow,
while the bottom line with $\bh$ is the pathway for the fine-grained task-irrelevant information flow.

The fine-grained predictor $\Theta^+_{\text{p}}$ spits out prediction based on $\bz$, with a cross-entropy loss: 

\begin{equation}
\begin{aligned}
    &\mL_{\text{fine}}(\Theta_{\text{enc}}, \Theta^+_{\text{p}}) \\ = 
    &\sum_{(\bx, \bt_i, y, y_i) \in D'} [-\log P_{\Theta^+_{\text{p}}}(y_i|\bz)],
    \label{DPL-finer}
\end{aligned}
\end{equation}


Another crucial design in the DPL is that the concatenation of $\bh$ and $\bz$, $[\bh \circ \bz]$, is fed to decide the prediction of the sequence-level prediction:

\begin{equation}
\begin{aligned}
    &\mL_{\text{coarse}}(\Theta_{\text{enc}}, \Theta^*_{\text{p}}) \\
    = 
    &\sum_{(\bx, \bt_i, y, y_i) \in D'} [-\log P_{\Theta^*_{\text{p}}}(y|\bh \circ \bz)].
    \label{DPL-coarse}
\end{aligned}
\end{equation}

The gradient of this loss will update the model parameters on both pathways. To prevent the degenerated case where the two pathways act completely separately, we introduce another crucial part to DPL in the next subsection.

\subsubsection{Adversarial Training}
The current version of DPL could still work as two separate systems, which is deemed a degenerated case.
Therefore, to guarantee the mutual exclusiveness between the $\bh$ and the $\bz$, we introduce an adversarial training loss term to maximally reduce the fine-grained task-relevant information carried by $\bh$:

\begin{equation}
\begin{aligned}
    \mL_{\text{enc}}(\Theta_{\text{enc}}) = 
    \sum_{(\bx, \bt_i, y, y_i) \in D'} [-\log P_{\Theta^+_{\text{p}}}(1-y_i|\bh)],
    \label{eq:adv-1}
\end{aligned}
\end{equation}
\begin{equation}
    \mL_{\text{dis}}(\Theta^+_{\text{p}}) = 
    \sum_{(\bx, \bt_i, y, y_i) \in D'} [-\log P_{\Theta^+_{\text{p}}}(y_i|\bh)],
    \label{eq:adv-2}
\end{equation}
\begin{equation}
    \mL_{\text{adv}}(\Theta_{\text{enc}},\Theta^+_{\text{p}}) = \mL_{\text{dis}}(\Theta^+_{\text{p}}) + \lambda \mL_{\text{enc}}(\Theta_{\text{enc}}),
    \label{eq:adv-3}
\end{equation}
where $\lambda$ weighs the trade-off between $\Theta_{\text{enc}}$ and $\Theta^+_{\text{p}}$.
The adversarial training was first introduced in \citet{lample2017fader} and has been widely used ~\cite{zhao2018adversarially,fu2018style,shen2017style,melnyk2017improved}.
The loss term trains $\Theta_{\text{enc}}$ to fool $\Theta^+_{\text{p}}$ by removing fine-grained task relevant information from $\bh$.
Considering that $\bz$ is only required by the fine-grained task, the less fine-grained task-relevant information the $\bh$ has, the less overlap there is between the $\bh$ and $\bz$.
As a result, the adversarial training makes h and z more mutually exclusive in terms of the carried information.

\subsubsection{Loss Function}
The overall loss function to optimzie DPL combines as below:
\begin{equation}
\begin{aligned}
    \mL(\Theta_{\text{enc}}, \Theta^*_{\text{p}}, \Theta^+_{\text{p}}) = &\mL_{\text{fine}}(\Theta_{\text{enc}}, \Theta^+_{\text{p}})\\ + 
    &\alpha \mL_{\text{coarse}}(\Theta_{\text{enc}}, \Theta^*_{\text{p}})\\ + 
    &\beta \mL_{\text{adv}}(\Theta_{\text{enc}}, \Theta^+_{\text{p}})
\end{aligned}
\end{equation}
where $\alpha$ and $\beta$ are weighing terms.
With this design of the loss functions, we posit that all three philosophies should be satisfied. The ideal result for it is that (i)-$\bz$ only carries information dedicated at the fine-level; (ii)-$\bh$ carries the information of the entire coarse level (i.e., the whole sequence) excluding the information of $\bz$; (iii)-neither $\bh$ nor $\bz$ is sufficient on deciding the whole-sequence coarse-level prediction, but with the concatenation of them, $\bh \circ \bz$, the information is just adequate.

\subsection{Grounding DPL to ABSA}


\subsubsection{Document-level Sentiment Analysis.}
The task aims to analyze the sentiments reflected by sentences.
Given an ordinary labeled document-level dataset $D = \{(\bx^0, y^0), (\bx^1, y^1) \dots (\bx^N, y^N)\}$ , where $\bx^i$ donates a sentence and $y^i$ donates the sentiment polarity of the sentence.
The goal of the task is to learn a mapping function: $f_{\text{sent}}(\bx^i) \to y^i$.

\subsubsection{Aspect-based Sentiment Analysis.}
The ABSA task is to derive the sentiment polarity attached to specific aspect terms in the given sentence.
Formally, one can draw a data point $(\bx^i, \by^i)$ from the dataset $D$. We assign a separate variable indicating the aspect terms annotation, $\{\bt^{i, 1}, \dots, \bt^{i, N_{i}}\}$, where $N_{i}$ denotes the number of total aspect terms in $\btau^i$.
In addition, the label $\by$ is a combination of polarities corresponding to aspect terms, $\by^i = \{y^{i, 1}, \dots, y^{i, N_{i}}\}$.
The goal for the ABSA is to learn the mapping $f_{\text{aspect}}(\bx^i, \bt^{i,k}) \to y^{i,k}$, where $k \in \{1,\dots, N_i\}$.

\subsubsection{Implementation}
Before implementing a specific DPL model, we first map the task objectives of the SA and ABSA tasks to the coarse- and fine-grained tasks in the DPL framework.
The coarse-grained task is the SA task, while the fine-grained task is the ABSA task.
In another word, the mapping $f_{\text{sent}}(\bx^i) \to y^i$, is considered as the coarse-grained mapping  $f_{\text{coarse}}(\bx) \to y$, and the mapping $f_{\text{aspect}}(\bx^i, \bt^{i,k}) \to y^{i,k}$ is considered as $f_{\text{fine}}(\bx, \bt_i) \to y_i$.

Then we choose the model for $\Theta_{\text{enc}}$, $\Theta^+_{\text{p}}$ and $\Theta^*_{\text{p}}$.
$\Theta^+_{\text{p}}$ and $\Theta^*_{\text{p}}$ are simple multilayer perceptron (MLP).
It is worth noting that $\Theta_{\text{enc}}$ can be a prior ABSA model.
Thus, we argue that the DPL framework can be applied to most ABSA methods.
Typically, we choose ~\citet{bai2020investigating}'s and ~\citet{rietzler2019adapt}'s works and a multi-task learning baseline as examples to verify. The results are shown in Table ~\ref{X+DPL}.

\begin{table}
\centering
\resizebox{!}{28pt}{
\begin{tabular}{cccccccc}
\toprule
\multirow{2}{*}{\textbf{Dataset}} & 
\multicolumn{2}{c}{\textbf{positive}s} & 
\multicolumn{2}{c}{\textbf{neutral}} & 
\multicolumn{2}{c}{\textbf{negative}} \\
\cmidrule(lr){2-3} \cmidrule(lr){4-5} \cmidrule(lr){6-7}
&Train & Test & Train & Test & Train & Test \\

\midrule
Rest & 2164 & 727 & 637 & 196 & 807 & 196 \\
Laptop & 976 & 337 & 455 & 128 & 851 & 167 \\
\hline
\end{tabular}}
\caption{\label{Statistics-dataset} Statistics of SemEval 2014 task 4 subtask 2.}
\end{table}

\section{Experiments}


\subsection{Experimental Setup}

\subsubsection{Dataset}
The experiments of the DPL framework require at least two datasets at different granularities.
For the ABSA task, we select the SemEval dataset ~\cite{pontikisemeval} as the fine-grained sentiment task dataset and the Amazon reviews dataset from Kaggle\footnote{\url{www.kaggle.com/bittlingmayer/amazonreviews}} as the coarse-grained sentiment task dataset. The SemEval datasets are used as our core task dataset, and the Amazon reviews dataset is used as an auxiliary dataset.




\textbf{Dataset SemEval.}
This dataset is SemEval 2014 task 4 subtask2 ~\cite{pontikisemeval}.
It has two sub-datasets, the reviews in the restaurant and laptop domains.
We show more details in Table~\ref{Statistics-dataset}. 

\textbf{Dataset Amazon Reviews.}
The dataset contains 3.6 million sentences in the training set and 0.4 million sentences in the test set. Considering the huge data volume gap, we only chose the test set as the auxiliary dataset for this experiment. 

\subsubsection{Generation of Pseudo Labels}
Here we provide some details of the pseudo labels generation process.


As a result of the PL generation, the ABSA dataset has true aspect-level sentimental labels and pseudo-sentence-level sentimental labels, while the SA dataset has true sentence-level sentimental labels and pseudo-aspect-level sentimental labels.

To get aspect terms from the sentence in the SA dataset, we first performed aspect extraction using the model proposed by ~\citet{li-etal-2019-exploiting} and discarded sentences without aspect terms.

We train the model proposed by ~\cite{bai2020investigating} as the teacher models on the aspect-level dataset with the accuracy scores of 86.05\% and 79.53\% respectively on the domain of Restaurant and Laptop.

We train a BERT+Linear as the teacher model on the document-level dataset, with a 94.45\% accuracy score in the restaurant Domain and a 93.35\% accuracy score in the laptop domain.





\subsubsection{Implementation Details}
In addition to the above introduction, some more important details of our experiments need to be clarified for ease of understanding.

\textbf{Evaluate Matrix}

The model for \textbf{ABSA} is tested on SemEval's test set. Like those who have performed this work before, we use the model classification accuracy (ACC) and macro-F1 (F1) scores as the evaluation criterion.


\textbf{Batch Loader}

Since the size of the current auxiliary dataset is much larger than the existing dataset. To avoid the large auxiliary dataset changing the original dataset distribution, we adopt two asynchronous loaders and define the step ratio $k$, i.e., whenever the model is trained on the original dataset by $1$ step, it is trained on the auxiliary dataset by $k$ steps. In general, we set $k=1$.

\textbf{Model Implementation}


The encoder has three main structures for the \textbf{ABSA} task: BERT~\cite{devlin2018bert}, Relational Graph Attention Networks (RGAT) ~\cite{wang2020relational}, and masking embedding module. 
The BERT and RGAT have been proved to have a good effect on this task.
The mask embedding module is used to generate $\bz$ and $\bh$. It is similar to the implementation of ``segment\_id'' in the code of BERT. 

\subsection{Main Results}
\begin{table}
\centering
\resizebox{!}{60pt}{
\begin{tabular}{cc|p{1cm}p{1cm}p{1cm}p{1cm}}
\toprule
\multicolumn{2}{c}{\multirow{2}{*}{\textbf{Model}}} & \multicolumn{2}{c}{\textbf{Restaurant}} & \multicolumn{2}{c}{\textbf{Laptop}}\\
\cmidrule(lr){3-4} \cmidrule(lr){5-6}
\multicolumn{2}{c}{}& \centering Acc &\centering F1 &\centering Acc &\centering F1 
\tabularnewline
\midrule
\multirow{4}{*}{\textbf{Auxiliary}}
&\textbf{~\citet{he2018exploiting}} &\centering 78.73 &\centering 68.63 &\centering 71.91 &\centering 68.79
\tabularnewline
&\textbf{~\citet{chen2019transfer}} &\centering 79.55 &\centering 71.41 &\centering 73.87 &\centering 70.10
\tabularnewline
&\textbf{~\citet{he2019interactive}} &\centering 83.89 &\centering 75.66 &\centering 75.36 &\centering 72.02
\tabularnewline
&\textbf{~\citet{liang2020iterative}} &\centering 84.93 &\centering 76.66 & \centering77.51 &\centering 73.42
\tabularnewline
\midrule
\multirow{4}{*}{\textbf{BERT}}
&\textbf{~\citet{bai2020investigating}*} &\centering 86.04 &\centering 80.27 &\centering 79.53 & \centering74.54
\tabularnewline
&\textbf{~\citet{pang2021dynamic}} &\centering 87.66 &\centering 82.97 &\centering 80.22 & \centering 77.28 
\tabularnewline
&\textbf{~\citet{li2021dual}} &\centering 87.13 &\centering 81.16 &\centering 81.80 & \centering78.10
\tabularnewline
&\textbf{\citet{rietzler2019adapt}} &\centering 87.89 &\centering 81.05 & \centering 80.23 &\centering 75.77
\tabularnewline
\midrule
\multirow{1}{*}{\textbf{Ours}}
& \textbf{DPL} &\centering \textbf{89.54} &\centering \textbf{84.86} & \centering \textbf{81.96} & \centering \textbf{78.58}
\tabularnewline
\hline
\end{tabular}
}
\caption{\label{main-result} Results of different methods. ``BERT'' represents the works that are also based on the BERT ~\cite{devlin2018bert}, ``Auxiliary'' represents the methods that also utilize auxiliary datasets to help the ABSA task. ``*'' means our replication results. The results show that our method achieves state-of-the-art in this benchmark.}
\end{table}
Table~\ref{main-result} shows that the DPL has achieved a state-of-the-art (SOTA) performance in terms of the average accuracy and F1-scores on the SemEval 2014 task 4 subtask 2 dataset.
The group denoted as ``Auxiliary Dataset is multi-task learning methods based on labeled datasets. Compared with them, our work shows the advantage of the PL method. 
``BERT-based'' are some recently published works with good results. Obviously, our method achieves significant improvements over them.

It should be noted that our design is based on the BERT. 
Thus the comparison is not made with the methods based on a more powerful pre-trained model, such as Roberta ~\cite{liu2019roberta}, DeBERTa ~\cite{silvaaspect}, and GPT-3 ~\cite{floridi2020gpt}.





\subsection{DPL as a General Framework}
\begin{table}
\centering
\resizebox{!}{60pt}{
\begin{tabular}{c|p{1cm}p{1cm}p{1cm}p{1cm}}
\toprule
\multirow{2}{*}{\textbf{Model}} & \multicolumn{2}{c}{\textbf{Restaurant}}  &\multicolumn{2}{c}{\textbf{Laptop}}\\
\cmidrule(lr){2-3} \cmidrule(lr){4-5} 
& \centering Acc &\centering  F1 &\centering  Acc &\centering  F1
\tabularnewline
\midrule
\textbf{RGAT~\cite{bai2020investigating}} &\centering 86.04 &\centering 80.27 &\centering 79.53 & \centering74.54
\tabularnewline
\textbf{RGAT+DPL} &\centering 87.22 &\centering 81.47 & \centering 81.01 & \centering 77.52
\tabularnewline
\textbf{Improvement} &\centering \textbf{+1.18} &\centering \textbf{+1.20} & \centering \textbf{+1.48} & \centering \textbf{+2.98}
\tabularnewline
\midrule
\textbf{Adapter\cite{rietzler2019adapt}} &\centering 87.89 &\centering 81.05 & \centering 80.23 & \centering 75.77
\tabularnewline
\textbf{Adapter+DPL} & \centering 89.54 &\centering 84.86 & \centering 81.96 & \centering 78.58
\tabularnewline
\textbf{Improvement} &\centering \textbf{+1.65} &\centering \textbf{+3.71} & \centering \textbf{+1.73} & \centering \textbf{+2.81}
\tabularnewline
\midrule
\textbf{MultiBERT} &\centering 84.54 &\centering 78.52 & \centering 78.32 & \centering 73.87
\tabularnewline
\textbf{MultiBERT+DPL} & \centering 85.52 &\centering 79.61 & \centering 79.75 & \centering 75.80
\tabularnewline
\textbf{Improvement} &\centering \textbf{+0.98} &\centering \textbf{+1.09} & \centering \textbf{+1.43} & \centering \textbf{+1.93}
\tabularnewline
\hline
\end{tabular}
}
\caption{Results of Combining DPL with Other Methods. Restaurant and Laptop are two benchmarks same as those in Table~\ref{main-result}. RGAT ~\cite{bai2020investigating}, Adapter ~\cite{rietzler2019adapt} are typical ABSA methods.
MultiBERT is a multi-task baseline implemented by us. 
It predicts the SA label based on the ``[cls]'' and predicts the ABSA task based on the specific word vector.
We add the DPL framework to them, denoted as ``+DPL'', and achieve significant improvements.
}
\label{X+DPL}
\end{table}
As we mentioned, we promote DPL as a general framework capable of combining other methods on the ABSA task. Table ~\ref{X+DPL} shows the performances of some typical methods before and after they combine the DPL framework. 
On the one hand, RGAT ~\cite{bai2020investigating} is a model architecture based on GAT and BERT. Thus the improvement shows that the DPL framework fits other architectural designs, even without auxiliary datasets.
On the other hand, for those methods involving auxiliary datasets, we take Adapter ~\cite{rietzler2019adapt} and MultiBERT for demonstration. 
Previous works are mainly divided into two categories, pretraining and multi-task learning. Adapter ~\cite{rietzler2019adapt} can be categorized into the pretraining class while MultiBERT is a multi-task learning baseline inspired by ~\citet{he2018exploiting}.
Since the previous works using the multi-task method to combine the SA and the ABSA datasets were LSTM based, we implemented a better model based on the BERT.
All the improvements verify that the DPL framework does not conflict with these methods and exhibits full compatibility for further performance gains.

\subsection{Ablation Study}
\begin{table}
\centering
\resizebox{!}{43pt}{
\begin{tabular}{c|p{1cm}p{1cm}p{1cm}p{1cm}}
\toprule
\multirow{2}{*}{\textbf{Model}} & \multicolumn{2}{c}{\textbf{Restaurant+Pre}}  &\multicolumn{2}{c}{\textbf{Restaurant}}\\
\cmidrule(lr){2-3} \cmidrule(lr){4-5} 
& \centering Acc &\centering  F1 &\centering  Acc &\centering  F1
\tabularnewline
\midrule
\textbf{DPL} & \centering 89.54 &\centering 84.86 &\centering86.68&\centering 80.44
\tabularnewline
\midrule
\textbf{Traditional Pseudo-Label} &\centering -1.43 &\centering -2.09 &\centering -1.60 &\centering -2.73
\tabularnewline
\midrule
\textbf{- adversarial training} &\centering -1.96 &\centering -3.31 &\centering -1.96 & \centering -3.60
\tabularnewline
\textbf{- coarse-grained pseudo labels } &\centering -1.60 &\centering -2.74  &\centering -1.34 &\centering -1.35 
\tabularnewline
\textbf{- fine-grained pseudo labels } &\centering -1.96 &\centering -2.84 &\centering -0.79 &\centering -1.79 
\tabularnewline
\hline
\end{tabular}
}
\caption{ Results of ablation study. ``Restaurant'' takes plain BERT as the initial model while ``Restaurant+Pre'' takes \citet{rietzler2019adapt}'s BERT as the initial model. ``DPL'' denotes our method. 
``Traditional Pseudo-Label'' represents we take the PL method for fine-grained tasks dropped out the coarse-grained labels. 
The last three cases named in the form of ``- X'' means that we deleted the ``X'' from the original DPL to evaluate the effect of ``X''.
}
\label{ablation}
\end{table}

We set up several sets of ablation experiments and present the results in Table ~\ref{ablation} to explore the role of adversarial training and pseudo labels in the DPL framework.

The above experiments contain two types of BERT on the SemEval Restaurant dataset. To ensure the fairness of the ablation experiments, we use the same parameters when training the same group, and the parameter configurations are shown in Appendix.

The comparison with ``Traditional Pseudo-Label'' shows the advantages of our method.
From the item ``- adversarial training'', the significant decline on F1 reflects that adversarial training plays an important role in the DPL framework. 
The items, ``- coarse-grained pseudo labels'' and ``- fine-grained pseudo labels'', show that only adding adversarial training at one granularity has less effect than adding it both ways.

Furthermore, we also take Chamfer Distance (CD) between the set of $\bh$ and the set of $\bz$ to provide an insight into the effect of the mutual exclusiveness. 
And the CD of the model with the adversarial training process is 30\% larger than that of the model without this process.
That means the adversarial training process increases the distance between the variable $\bh$ and $\bz$.

\section{Conclusion}
In this paper, we propose Dual-granularity Pseudo Labeling (DPL). DPL extends from the vanilla Pseudo-Label method and augments it to a dual-pathway system. It additionally enforces strong control of information flow directing to the data at different granularities of annotation. The results demonstrate the state-of-the-art performance of DPL on the data-scarce ABSA task. As a pioneering framework design, we also show that the DPL is compatible with pre-training and multi-task learning methods as published before. In the future, we hope to explore the possibility of DPL in other domains, such as computer vision problems where the discrepancy of granularities possesses.
\bibliography{acl}

\begin{thebibliography}{52}
\expandafter\ifx\csname natexlab\endcsname\relax\def\natexlab#1{#1}\fi

\bibitem[{Bai et~al.(2020)Bai, Liu, and Zhang}]{bai2020investigating}
Xuefeng Bai, Pengbo Liu, and Yue Zhang. 2020.
\newblock Investigating typed syntactic dependencies for targeted sentiment
  classification using graph attention neural network.
\newblock \emph{IEEE/ACM Transactions on Audio, Speech, and Language
  Processing}.

\bibitem[{Chen and Qian(2019)}]{chen2019transfer}
Zhuang Chen and Tieyun Qian. 2019.
\newblock Transfer capsule network for aspect level sentiment classification.
\newblock In \emph{Proceedings of the 57th Annual Meeting of the Association
  for Computational Linguistics}, pages 547--556.

\bibitem[{Devlin et~al.(2018)Devlin, Chang, Lee, and
  Toutanova}]{devlin2018bert}
Jacob Devlin, Ming-Wei Chang, Kenton Lee, and Kristina Toutanova. 2018.
\newblock Bert: Pre-training of deep bidirectional transformers for language
  understanding.
\newblock \emph{arXiv preprint arXiv:1810.04805}.

\bibitem[{Dong et~al.(2014)Dong, Wei, Tan, Tang, Zhou, and
  Xu}]{dong2014adaptive}
Li~Dong, Furu Wei, Chuanqi Tan, Duyu Tang, Ming Zhou, and Ke~Xu. 2014.
\newblock Adaptive recursive neural network for target-dependent twitter
  sentiment classification.
\newblock In \emph{Proceedings of the 52nd annual meeting of the association
  for computational linguistics (volume 2: Short papers)}, pages 49--54.

\bibitem[{Floridi and Chiriatti(2020)}]{floridi2020gpt}
Luciano Floridi and Massimo Chiriatti. 2020.
\newblock Gpt-3: Its nature, scope, limits, and consequences.
\newblock \emph{Minds and Machines}, 30(4):681--694.

\bibitem[{Fu et~al.(2018)Fu, Tan, Peng, Zhao, and Yan}]{fu2018style}
Zhenxin Fu, Xiaoye Tan, Nanyun Peng, Dongyan Zhao, and Rui Yan. 2018.
\newblock Style transfer in text: Exploration and evaluation.
\newblock In \emph{Proceedings of the AAAI Conference on Artificial
  Intelligence}, volume~32.

\bibitem[{Gao et~al.(2019)Gao, Feng, Song, and Wu}]{gao2019target}
Zhengjie Gao, Ao~Feng, Xinyu Song, and Xi~Wu. 2019.
\newblock Target-dependent sentiment classification with bert.
\newblock \emph{IEEE Access}, 7:154290--154299.

\bibitem[{Ge et~al.(2020)Ge, Chen, and Li}]{ge2020mutual}
Yixiao Ge, Dapeng Chen, and Hongsheng Li. 2020.
\newblock Mutual mean-teaching: Pseudo label refinery for unsupervised domain
  adaptation on person re-identification.
\newblock \emph{arXiv preprint arXiv:2001.01526}.

\bibitem[{He et~al.(2019{\natexlab{a}})He, Gu, Shen, and
  Ranzato}]{he2019revisiting}
Junxian He, Jiatao Gu, Jiajun Shen, and Marc'Aurelio Ranzato.
  2019{\natexlab{a}}.
\newblock Revisiting self-training for neural sequence generation.
\newblock \emph{arXiv preprint arXiv:1909.13788}.

\bibitem[{He et~al.(2018)He, Lee, Ng, and Dahlmeier}]{he2018exploiting}
Ruidan He, Wee~Sun Lee, Hwee~Tou Ng, and Daniel Dahlmeier. 2018.
\newblock Exploiting document knowledge for aspect-level sentiment
  classification.
\newblock \emph{arXiv preprint arXiv:1806.04346}.

\bibitem[{He et~al.(2019{\natexlab{b}})He, Lee, Ng, and
  Dahlmeier}]{he2019interactive}
Ruidan He, Wee~Sun Lee, Hwee~Tou Ng, and Daniel Dahlmeier. 2019{\natexlab{b}}.
\newblock An interactive multi-task learning network for end-to-end
  aspect-based sentiment analysis.
\newblock \emph{arXiv preprint arXiv:1906.06906}.

\bibitem[{Huang and Carley(2019)}]{huang2019syntax}
Binxuan Huang and Kathleen~M Carley. 2019.
\newblock Syntax-aware aspect level sentiment classification with graph
  attention networks.
\newblock \emph{arXiv preprint arXiv:1909.02606}.

\bibitem[{Huang et~al.(2018)Huang, Ou, and Carley}]{huang2018aspect}
Binxuan Huang, Yanglan Ou, and Kathleen~M Carley. 2018.
\newblock Aspect level sentiment classification with attention-over-attention
  neural networks.
\newblock In \emph{International Conference on Social Computing,
  Behavioral-Cultural Modeling and Prediction and Behavior Representation in
  Modeling and Simulation}, pages 197--206. Springer.

\bibitem[{Jiang et~al.(2011)Jiang, Yu, Zhou, Liu, and Zhao}]{jiang2011target}
Long Jiang, Mo~Yu, Ming Zhou, Xiaohua Liu, and Tiejun Zhao. 2011.
\newblock Target-dependent twitter sentiment classification.
\newblock In \emph{Proceedings of the 49th annual meeting of the association
  for computational linguistics: human language technologies}, pages 151--160.

\bibitem[{Lample et~al.(2017)Lample, Zeghidour, Usunier, Bordes, Denoyer, and
  Ranzato}]{lample2017fader}
Guillaume Lample, Neil Zeghidour, Nicolas Usunier, Antoine Bordes, Ludovic
  Denoyer, and Marc'Aurelio Ranzato. 2017.
\newblock Fader networks: Manipulating images by sliding attributes.
\newblock \emph{arXiv preprint arXiv:1706.00409}.

\bibitem[{Li et~al.(2018)Li, Liu, and Zhou}]{li2018hierarchical}
Lishuang Li, Yang Liu, and AnQiao Zhou. 2018.
\newblock Hierarchical attention based position-aware network for aspect-level
  sentiment analysis.
\newblock In \emph{Proceedings of the 22nd conference on computational natural
  language learning}, pages 181--189.

\bibitem[{Li et~al.(2021)Li, Chen, Feng, Ma, Wang, and Hovy}]{li2021dual}
Ruifan Li, Hao Chen, Fangxiang Feng, Zhanyu Ma, Xiaojie Wang, and Eduard Hovy.
  2021.
\newblock Dual graph convolutional networks for aspect-based sentiment
  analysis.
\newblock In \emph{Proceedings of the 59th Annual Meeting of the Association
  for Computational Linguistics and the 11th International Joint Conference on
  Natural Language Processing (Volume 1: Long Papers)}, pages 6319--6329.

\bibitem[{Li et~al.(2019)Li, Bing, Zhang, and Lam}]{li-etal-2019-exploiting}
Xin Li, Lidong Bing, Wenxuan Zhang, and Wai Lam. 2019.
\newblock \href {https://www.aclweb.org/anthology/D19-5505} {Exploiting {BERT}
  for end-to-end aspect-based sentiment analysis}.
\newblock In \emph{Proceedings of the 5th Workshop on Noisy User-generated Text
  (W-NUT 2019)}, pages 34--41.

\bibitem[{Liang et~al.(2020)Liang, Meng, Zhang, Xu, Chen, and
  Zhou}]{liang2020iterative}
Yunlong Liang, Fandong Meng, Jinchao Zhang, Jinan Xu, Yufeng Chen, and Jie
  Zhou. 2020.
\newblock An iterative knowledge transfer network with routing for aspect-based
  sentiment analysis.
\newblock \emph{arXiv preprint arXiv:2004.01935}.

\bibitem[{Liu et~al.(2021)Liu, Yuan, Fu, Jiang, Hayashi, and
  Neubig}]{liu2021pre}
Pengfei Liu, Weizhe Yuan, Jinlan Fu, Zhengbao Jiang, Hiroaki Hayashi, and
  Graham Neubig. 2021.
\newblock Pre-train, prompt, and predict: A systematic survey of prompting
  methods in natural language processing.
\newblock \emph{arXiv preprint arXiv:2107.13586}.

\bibitem[{Liu et~al.(2020)Liu, Li, Wu, Su, and Sun}]{liu2020jointly}
Shu Liu, Wei Li, Yunfang Wu, Qi~Su, and Xu~Sun. 2020.
\newblock Jointly modeling aspect and sentiment with dynamic heterogeneous
  graph neural networks.
\newblock \emph{arXiv preprint arXiv:2004.06427}.

\bibitem[{Liu et~al.(2019)Liu, Ott, Goyal, Du, Joshi, Chen, Levy, Lewis,
  Zettlemoyer, and Stoyanov}]{liu2019roberta}
Yinhan Liu, Myle Ott, Naman Goyal, Jingfei Du, Mandar Joshi, Danqi Chen, Omer
  Levy, Mike Lewis, Luke Zettlemoyer, and Veselin Stoyanov. 2019.
\newblock Roberta: A robustly optimized bert pretraining approach.
\newblock \emph{arXiv preprint arXiv:1907.11692}.

\bibitem[{Ma et~al.(2017)Ma, Li, Zhang, and Wang}]{ma2017interactive}
Dehong Ma, Sujian Li, Xiaodong Zhang, and Houfeng Wang. 2017.
\newblock Interactive attention networks for aspect-level sentiment
  classification.
\newblock \emph{arXiv preprint arXiv:1709.00893}.

\bibitem[{Mallis et~al.(2020)Mallis, Sanchez, Bell, and
  Tzimiropoulos}]{mallis2020unsupervised}
Dimitrios Mallis, Enrique Sanchez, Matthew Bell, and Georgios Tzimiropoulos.
  2020.
\newblock Unsupervised learning of object landmarks via self-training
  correspondence.
\newblock \emph{Advances in Neural Information Processing Systems}, 33.

\bibitem[{Melnyk et~al.(2017)Melnyk, Santos, Wadhawan, Padhi, and
  Kumar}]{melnyk2017improved}
Igor Melnyk, Cicero Nogueira~dos Santos, Kahini Wadhawan, Inkit Padhi, and
  Abhishek Kumar. 2017.
\newblock Improved neural text attribute transfer with non-parallel data.
\newblock \emph{arXiv preprint arXiv:1711.09395}.

\bibitem[{Mukherjee and Awadallah(2020)}]{mukherjee2020uncertainty}
Subhabrata Mukherjee and Ahmed Awadallah. 2020.
\newblock Uncertainty-aware self-training for few-shot text classification.
\newblock \emph{Advances in Neural Information Processing Systems}, 33.

\bibitem[{Nguyen and Shirai(2015)}]{nguyen2015phrasernn}
Thien~Hai Nguyen and Kiyoaki Shirai. 2015.
\newblock Phrasernn: Phrase recursive neural network for aspect-based sentiment
  analysis.
\newblock In \emph{Proceedings of the 2015 Conference on Empirical Methods in
  Natural Language Processing}, pages 2509--2514.

\bibitem[{Oh et~al.(2021)Oh, Lee, Whang, Park, Seo, Kim, and Kim}]{oh2021deep}
Shinhyeok Oh, Dongyub Lee, Taesun Whang, IlNam Park, Gaeun Seo, EungGyun Kim,
  and Harksoo Kim. 2021.
\newblock Deep context-and relation-aware learning for aspect-based sentiment
  analysis.
\newblock \emph{arXiv preprint arXiv:2106.03806}.

\bibitem[{Pang et~al.(2021)Pang, Xue, Yan, Huang, and Feng}]{pang2021dynamic}
Shiguan Pang, Yun Xue, Zehao Yan, Weihao Huang, and Jinhui Feng. 2021.
\newblock Dynamic and multi-channel graph convolutional networks for
  aspect-based sentiment analysis.
\newblock In \emph{Findings of the Association for Computational Linguistics:
  ACL-IJCNLP 2021}, pages 2627--2636.

\bibitem[{Pham et~al.(2020)Pham, Dai, Xie, Luong, and Le}]{pham2020meta}
Hieu Pham, Zihang Dai, Qizhe Xie, Minh-Thang Luong, and Quoc~V Le. 2020.
\newblock Meta pseudo labels.
\newblock \emph{arXiv preprint arXiv:2003.10580}.

\bibitem[{Pontiki et~al.()Pontiki, Papageorgiou, Galanis, Androutsopoulos,
  Pavlopoulos, and Manandhar}]{pontikisemeval}
Maria Pontiki, Haris Papageorgiou, Dimitrios Galanis, Ion Androutsopoulos, John
  Pavlopoulos, and Suresh Manandhar.
\newblock Semeval-2014 task 4: Aspect based sentiment analysis.

\bibitem[{Rietzler et~al.(2019)Rietzler, Stabinger, Opitz, and
  Engl}]{rietzler2019adapt}
Alexander Rietzler, Sebastian Stabinger, Paul Opitz, and Stefan Engl. 2019.
\newblock Adapt or get left behind: Domain adaptation through bert language
  model finetuning for aspect-target sentiment classification.
\newblock \emph{arXiv preprint arXiv:1908.11860}.

\bibitem[{Shen et~al.(2017)Shen, Lei, Barzilay, and Jaakkola}]{shen2017style}
Tianxiao Shen, Tao Lei, Regina Barzilay, and Tommi Jaakkola. 2017.
\newblock Style transfer from non-parallel text by cross-alignment.
\newblock \emph{arXiv preprint arXiv:1705.09655}.

\bibitem[{Silva and Marcacini()}]{silvaaspect}
Emanuel~H Silva and Ricardo~M Marcacini.
\newblock Aspect-based sentiment analysis using bert with disentangled
  attention.

\bibitem[{Song et~al.(2019)Song, Wang, Jiang, Liu, and
  Rao}]{song2019attentional}
Youwei Song, Jiahai Wang, Tao Jiang, Zhiyue Liu, and Yanghui Rao. 2019.
\newblock Attentional encoder network for targeted sentiment classification.
\newblock \emph{arXiv preprint arXiv:1902.09314}.

\bibitem[{Sun et~al.(2019)Sun, Zhang, Mensah, Mao, and Liu}]{sun2019aspect}
Kai Sun, Richong Zhang, Samuel Mensah, Yongyi Mao, and Xudong Liu. 2019.
\newblock Aspect-level sentiment analysis via convolution over dependency tree.
\newblock In \emph{Proceedings of the 2019 Conference on Empirical Methods in
  Natural Language Processing and the 9th International Joint Conference on
  Natural Language Processing (EMNLP-IJCNLP)}, pages 5683--5692.

\bibitem[{Tai et~al.(2015)Tai, Socher, and Manning}]{tai2015improved}
Kai~Sheng Tai, Richard Socher, and Christopher~D Manning. 2015.
\newblock Improved semantic representations from tree-structured long
  short-term memory networks.
\newblock \emph{arXiv preprint arXiv:1503.00075}.

\bibitem[{Tang et~al.(2016)Tang, Qin, and Liu}]{tang2016aspect}
Duyu Tang, Bing Qin, and Ting Liu. 2016.
\newblock Aspect level sentiment classification with deep memory network.
\newblock \emph{arXiv preprint arXiv:1605.08900}.

\bibitem[{Vo and Zhang(2015)}]{vo2015target}
Duy-Tin Vo and Yue Zhang. 2015.
\newblock Target-dependent twitter sentiment classification with rich automatic
  features.
\newblock In \emph{Twenty-fourth international joint conference on artificial
  intelligence}.

\bibitem[{Wang et~al.(2020)Wang, Shen, Yang, Quan, and
  Wang}]{wang2020relational}
Kai Wang, Weizhou Shen, Yunyi Yang, Xiaojun Quan, and Rui Wang. 2020.
\newblock Relational graph attention network for aspect-based sentiment
  analysis.
\newblock \emph{arXiv preprint arXiv:2004.12362}.

\bibitem[{Wang et~al.(2018)Wang, Mazumder, Liu, Zhou, and
  Chang}]{wang2018target}
Shuai Wang, Sahisnu Mazumder, Bing Liu, Mianwei Zhou, and Yi~Chang. 2018.
\newblock Target-sensitive memory networks for aspect sentiment classification.
\newblock In \emph{Proceedings of the 56th Annual Meeting of the Association
  for Computational Linguistics (Volume 1: Long Papers)}, pages 957--967.

\bibitem[{Wang et~al.(2016)Wang, Huang, Zhu, and Zhao}]{wang2016attention}
Yequan Wang, Minlie Huang, Xiaoyan Zhu, and Li~Zhao. 2016.
\newblock Attention-based lstm for aspect-level sentiment classification.
\newblock In \emph{Proceedings of the 2016 conference on empirical methods in
  natural language processing}, pages 606--615.

\bibitem[{Xie et~al.(2020)Xie, Luong, Hovy, and Le}]{xie2020self}
Qizhe Xie, Minh-Thang Luong, Eduard Hovy, and Quoc~V Le. 2020.
\newblock Self-training with noisy student improves imagenet classification.
\newblock In \emph{Proceedings of the IEEE/CVF Conference on Computer Vision
  and Pattern Recognition}, pages 10687--10698.

\bibitem[{Xu et~al.(2019)Xu, Liu, Shu, and Yu}]{xu2019bert}
Hu~Xu, Bing Liu, Lei Shu, and Philip~S Yu. 2019.
\newblock Bert post-training for review reading comprehension and aspect-based
  sentiment analysis.
\newblock \emph{arXiv preprint arXiv:1904.02232}.

\bibitem[{Yan et~al.(2021)Yan, Dai, Qiu, Zhang et~al.}]{yan2021unified}
Hang Yan, Junqi Dai, Xipeng Qiu, Zheng Zhang, et~al. 2021.
\newblock A unified generative framework for aspect-based sentiment analysis.
\newblock \emph{arXiv preprint arXiv:2106.04300}.

\bibitem[{Yang et~al.(2019)Yang, Zeng, Yang, Song, and Xu}]{yang2019multi}
Heng Yang, Biqing Zeng, JianHao Yang, Youwei Song, and Ruyang Xu. 2019.
\newblock A multi-task learning model for chinese-oriented aspect polarity
  classification and aspect term extraction.
\newblock \emph{Neurocomputing}, 419:344--356.

\bibitem[{Yu et~al.(2021{\natexlab{a}})Yu, Ao, Luo, Yang, Sun, Li, and
  He}]{yu2021making}
Guoxin Yu, Xiang Ao, Ling Luo, Min Yang, Xiaofei Sun, Jiwei Li, and Qing He.
  2021{\natexlab{a}}.
\newblock Making flexible use of subtasks: A multiplex interaction network for
  unified aspect-based sentiment analysis.
\newblock In \emph{Findings of the Association for Computational Linguistics:
  ACL-IJCNLP 2021}, pages 2695--2705.

\bibitem[{Yu et~al.(2021{\natexlab{b}})Yu, Gong, and Xia}]{yu2021cross}
Jianfei Yu, Chenggong Gong, and Rui Xia. 2021{\natexlab{b}}.
\newblock Cross-domain review generation for aspect-based sentiment analysis.
\newblock In \emph{Findings of the Association for Computational Linguistics:
  ACL-IJCNLP 2021}, pages 4767--4777.

\bibitem[{Zhang et~al.(2019)Zhang, Li, and Song}]{zhang2019aspect}
Chen Zhang, Qiuchi Li, and Dawei Song. 2019.
\newblock Aspect-based sentiment classification with aspect-specific graph
  convolutional networks.
\newblock \emph{arXiv preprint arXiv:1909.03477}.

\bibitem[{Zhang et~al.(2016)Zhang, Zhang, and Vo}]{zhang2016gated}
Meishan Zhang, Yue Zhang, and Duy-Tin Vo. 2016.
\newblock Gated neural networks for targeted sentiment analysis.
\newblock In \emph{Proceedings of the AAAI Conference on Artificial
  Intelligence}, volume~30.

\bibitem[{Zhao et~al.(2018)Zhao, Kim, Zhang, Rush, and
  LeCun}]{zhao2018adversarially}
Junbo Zhao, Yoon Kim, Kelly Zhang, Alexander Rush, and Yann LeCun. 2018.
\newblock Adversarially regularized autoencoders.
\newblock In \emph{International conference on machine learning}, pages
  5902--5911. PMLR.

\bibitem[{Zoph et~al.(2020)Zoph, Ghiasi, Lin, Cui, Liu, Cubuk, and
  Le}]{zoph2020rethinking}
Barret Zoph, Golnaz Ghiasi, Tsung-Yi Lin, Yin Cui, Hanxiao Liu, Ekin~D Cubuk,
  and Quoc~V Le. 2020.
\newblock Rethinking pre-training and self-training.
\newblock \emph{arXiv preprint arXiv:2006.06882}.

\end{thebibliography}
\bibliographystyle{acl_natbib}




\end{document}